\documentclass[runningheads]{llncs}
\usepackage{graphicx}
\usepackage{comment}
\usepackage{amsmath,amssymb} %
\usepackage{color}
\usepackage{multirow}
\usepackage[table,xcdraw]{xcolor}
\usepackage{bm}

\usepackage{ textcomp }

\usepackage{booktabs}
\usepackage{siunitx}

\usepackage{xspace}

\makeatletter
\DeclareRobustCommand\onedot{\futurelet\@let@token\@onedot}
\def\@onedot{\ifx\@let@token.\else.\null\fi\xspace}

\def\eg{\emph{e.g}\onedot} 
\def\ie{\emph{i.e}\onedot} 
 
\def\etc{\emph{etc}\onedot} 
 
\def\etal{\emph{et al}\onedot}
\makeatother

\begin{document}
\pagestyle{headings}
\mainmatter
\def\ECCVSubNumber{6826}  %
\def\XXX{\textcolor{red}{XXXX}}
\def\XXXX{\textcolor{red}{XXXX}}

\newcommand{\alphap}{\bm{\hat{\alpha}}}
\newcommand{\fp}{\mathbf{\hat{{F}}}}
\newcommand{\bp}{\mathbf{\hat{{B}}}}
\newcommand{\cgt}{\mathbf{C}}

\newcommand{\alphagt}{\bm{{\alpha}}}
\newcommand{\fgt}{\mathbf{{{F}}}}
\newcommand{\bgt}{\mathbf{{{B}}}}

\title{$F$, $B$, Alpha Matting } %

\titlerunning{FBA Matting}
\author{Marco Forte \and
Fran\c{c}ois Piti\'e} %
\institute{Trinity College Dublin
\email{\{fortem,fpitie\}@tcd.ie}}
\maketitle
\begin{abstract}
Cutting out an object and estimating its opacity mask, known as image matting, is a key task in many image editing applications. Deep learning approaches have made significant progress by adapting the encoder-decoder architecture of segmentation networks. However, most of the existing networks only predict the alpha matte and post-processing methods must then be used to recover the original foreground and background colours in the transparent regions. Recently, two methods have shown improved results by also estimating the foreground colours, but at a significant computational and memory cost.

In this paper, we propose a low-cost modification to alpha matting networks to also predict the foreground and background colours. We study variations of the training regime and explore a wide range of existing and novel loss functions for the joint prediction. 

Our method achieves the state of the art performance on the Adobe Composition-1k dataset for alpha matte and composite colour quality. It is also the current best performing method on the alphamatting.com online evaluation.

\end{abstract}

\section{Introduction}
\begin{figure*}[t]
 \centering
\setlength{\tabcolsep}{0.1em}
\begin{tabular}{cccc|cccc}
 & & &  &  & & &\\
\includegraphics[height=.12\linewidth,width=.16\linewidth]{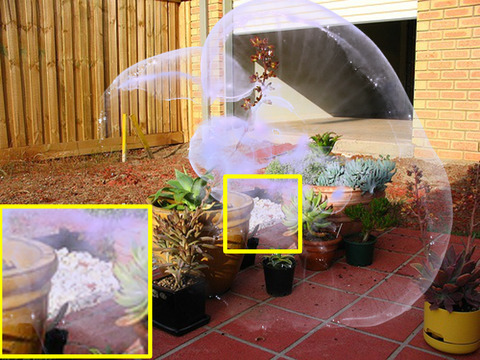}  &
\includegraphics[height=.12\linewidth,width=.16\linewidth]{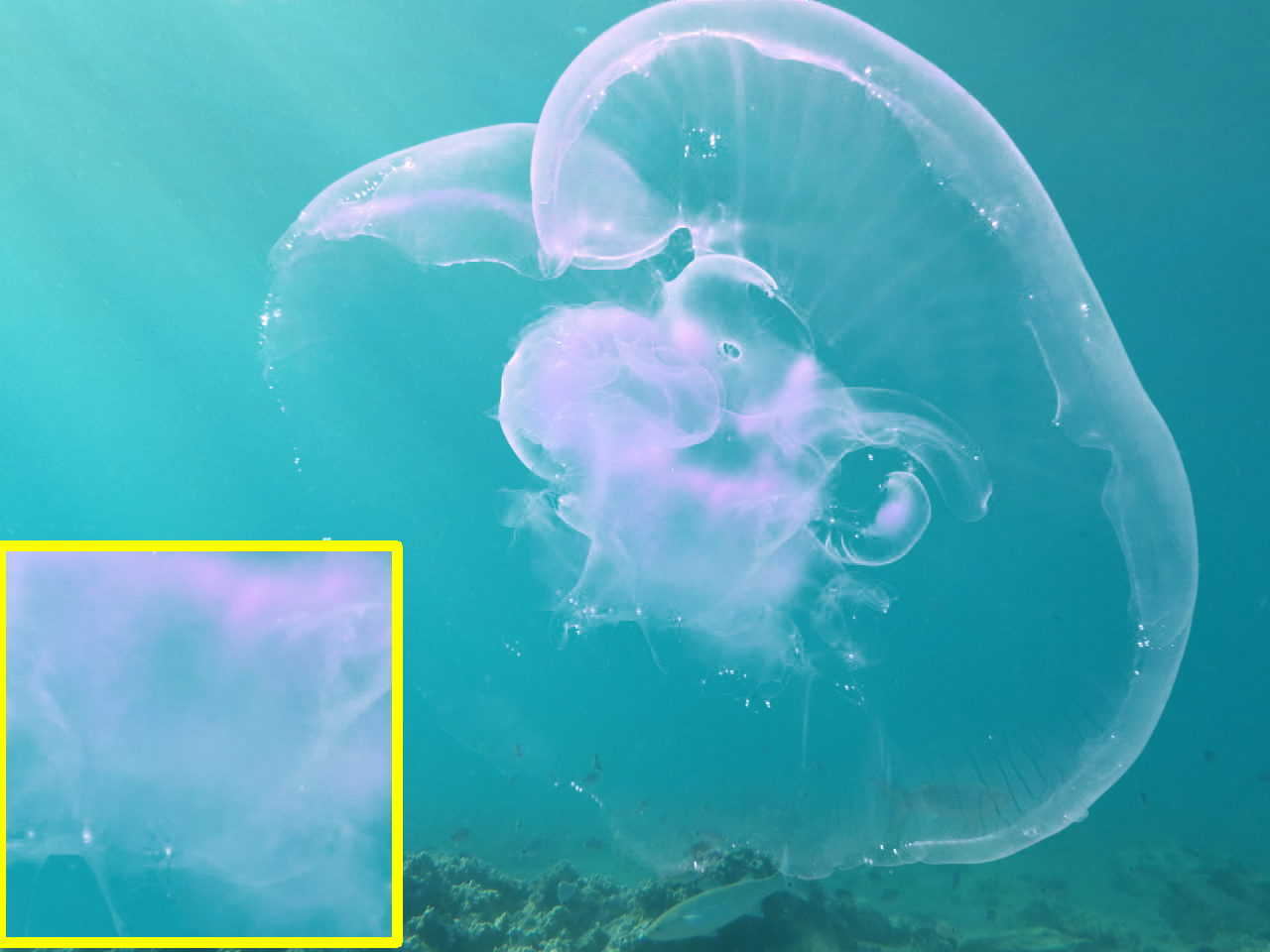}  &
\includegraphics[height=.12\linewidth,width=.16\linewidth]{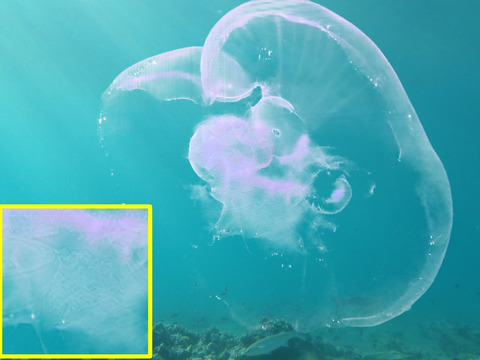}  & &  &
\includegraphics[height=.12\linewidth,width=.16\linewidth]{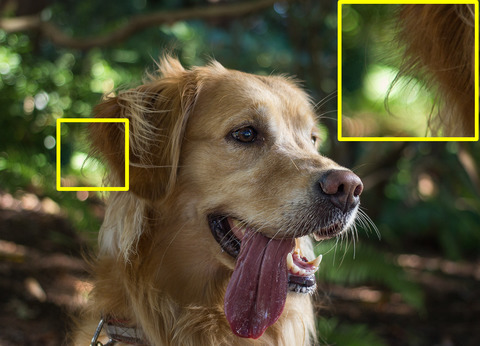}  &
\includegraphics[height=.12\linewidth,width=.16\linewidth]{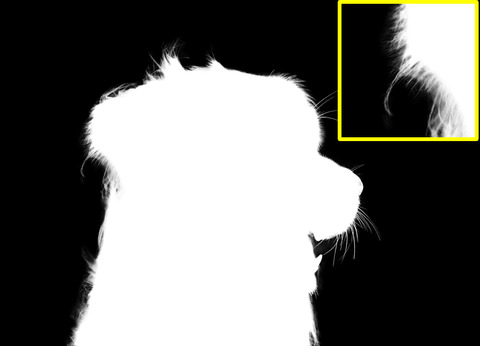}&
\includegraphics[height=.12\linewidth,width=.16\linewidth]{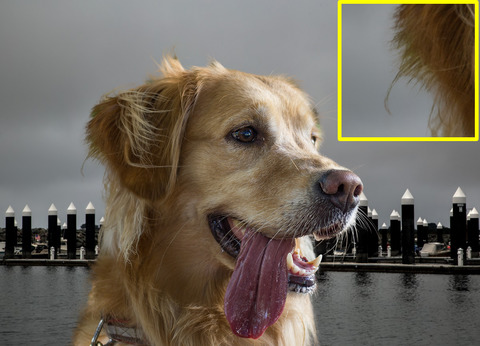}  \\
\tiny{Input Image}  & \tiny{GT:$\alpha_{gt}, F=F_{gt}$} & \tiny{Ours: $\alphap, F=\fp$}  &  & & \tiny{Input Image}  & \multicolumn{2}{c}{\tiny{Our alpha and composite}}\\
\includegraphics[height=.12\linewidth,width=.16\linewidth]{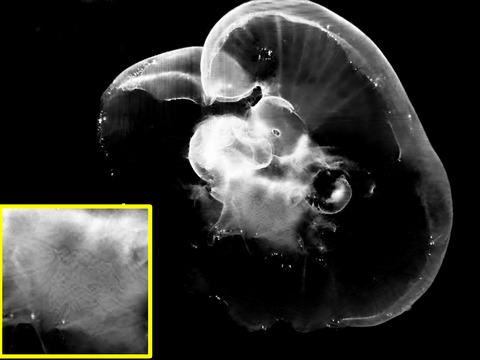}  &
\includegraphics[height=.12\linewidth,width=.16\linewidth]{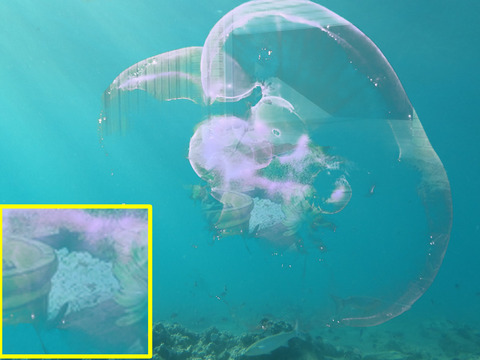}  &
\includegraphics[height=.12\linewidth,width=.16\linewidth]{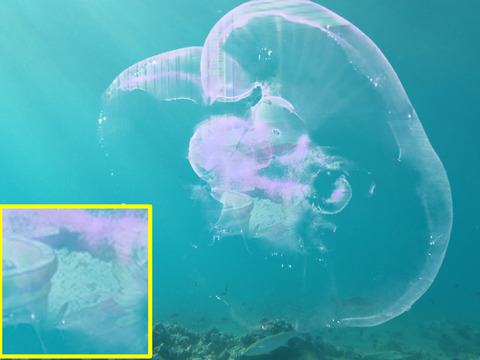}  &  &  &
\includegraphics[height=.12\linewidth,width=.16\linewidth]{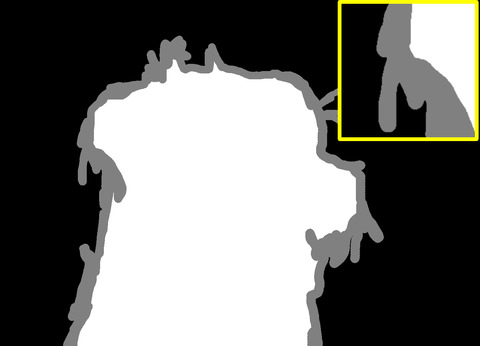} & 
\includegraphics[height=.12\linewidth,width=.16\linewidth]{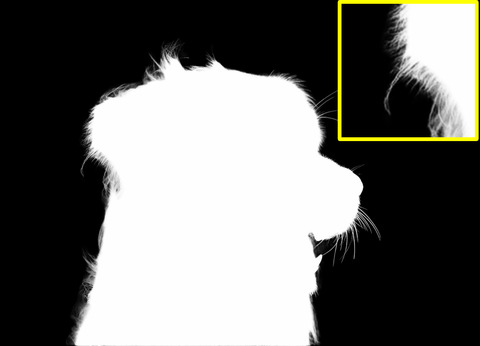}&
\includegraphics[height=.12\linewidth,width=.16\linewidth]{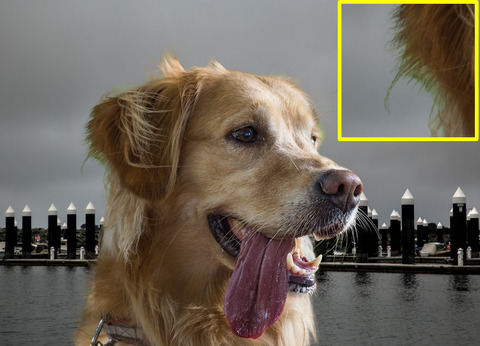}  \\
\tiny{Our alpha $\alphap$}  &\tiny{ Naive:  $\alphap, F=\alphap I$}  & \tiny{\cite{ClosedFormMattingPAMI}: $\alphap,F=F_{\text{\cite{ClosedFormMattingPAMI}}}$ }  &  & & \tiny{Trimap} & \multicolumn{2}{c}{\tiny{CA Matting~\cite{ContextMatting} alpha and composite}}\\

\end{tabular}
\caption{Alpha matting and compositing. Left: We show the need for foreground prediction and compare our composite with a post-processing method~\cite{ClosedFormMattingPAMI}. Right: We show our method better excludes background colours compared to the state of the art method for simultaneous prediction~\cite{ContextMatting}. }.\label{fig:realExamples}
\end{figure*}
Alpha matting refers to the problem of extracting the opacity mask, alpha matte,
of an object in an image. It is an essential task in visual media production as
it is required to selectively apply effects on image layers or recompose objects
onto new backgrounds. The alpha matting model for images is known as the
Compositing Equation~\cite{brinkman_compositing} and is defined as follows.
\begin{equation}
\cgt_i = \alphagt_i \fgt_i + (1-\alphagt_i)\bgt_i
\label{eq:matting}
\end{equation}
In this equation, $\cgt_i$ is the observed colour value at pixel site $i$,
$\fgt_i,\bgt_i$ are the pixel colours in the foreground and background layers at
that same site and $\alphagt_i$ is the level of mixing between the two
layers. An $\alpha$ value of 1 thus indicates a pure foreground pixel, a value
of 0 indicates a pure background pixel. The user would typically define a trimap
indicating regions of definite background ($\alpha=0$), definite foreground
($\alpha=1$) and unknown $\alpha$.

Estimating the $\alpha$-matte is clearly an ill-posed problem as we have 3
non-linear equations for 7 unknowns, without any well defined spatial dependency
to help us. The task is especially difficult when the foreground and background
colours are similar or when the background is highly textured but recently, deep
convolutional neural networks (CNNs) have progressed the state-of-the-art in
$\alpha$-matte prediction (\cite{DeepImageMatting,AlphaGAN,LearningBasedSamplingMatting,ContextMatting}) by better modelling natural image priors
and $\alpha$-mattes priors. These $\alpha$-matting networks are based on the
core architectures of semantic segmentation neural networks. This makes sense as
$\alpha$-matting can be seen as an extension of binary segmentation, where
instead of estimating a binary $\{0,1\}$ label field we estimate a floating
point $\alpha \in [0,1]$ field. For instance, the original \textit{Deep Image
  Matting} paper of ~\cite{DeepImageMatting} adapted the FCN architecture of
~\cite{FCN} to take as an input the original image and a trimap and as an output
the $\alpha$-matte instead of a binary mask.

As the level of details required in the output matte is much higher than in
segmentation, most of the recent literature has focused on making
architectural changes that can increase the resolution ability of the core
encoder-decoder architecture (see \cite{DeepImageMatting,IndexMatting,VDRNMatting,ContextMatting}). Curiously, most of these methods
only estimate $\alpha$, and not $\fgt$ or $\bgt$, which are however also required by
most applications. Un-mixing the foreground and background colours is usually
left as a post-processing step, (\eg Levin et al.\cite{ClosedFormMattingPAMI}). Recent works~\cite{ContextMatting,LearningBasedSamplingMatting}
have started to recognise the importance of jointly estimating $\alpha, F, B$,
but the resulting architectures are computationally expansive. In this paper, we
propose a novel architecture for jointly estimating $\alpha, F, B$ and
 study the benefit of  estimating $\fgt$ and $\bgt$ through the
examination of different possible loss functions.

Whereas in binary segmentation the choice of Loss functions is relatively straightforward (\ie
cross-entropy or IoU), the nature of the $\alpha$-matte gives us more options to consider. A few of the $\alpha$ losses proposed in previous
works include: Huber, $\mathcal{L}_1$, $\mathcal{L}_1$ on the gradient, pyramid Laplacian,
discriminative loss, \etc. Also predicting $\fgt$ and $\bgt$ further extends the
choice of possible loss functions and we propose to systematically study the
merits of these different losses.

Parallel to these considerations on loss and architecture, interesting issues
around training have also emerged. Indeed, as matting is an intrinsically harder
task than segmentation, training can be long and trickier to setup
correctly. Seemingly insignificant implementation details can turn out to be of
critical importance. A few research groups have reported difficulties in
reproducing the results of Deep Image Matting~\cite{foamliu_2020,AlphaGAN,joker316701882_2018}. We discovered that
simply setting the batch size to 1 in~\cite{DeepImageMatting} had a critical impact on the
performance and could, by itself, explain the reported training failures.

{\bf Contributions.} We propose therefore that, of equal importance to the core
architecture design, are the considerations around training and loss functions, and
the main contribution of this paper is to propose a novel architecture for
jointly estimating $\alpha, F, B$ and to systematically explore the impact of
key choices in losses and training regimes. Our study include 17 experiments
that contribute to three study areas:
\begin{enumerate}
    \item a comparison of min-batch and stochastic gradient descent and the use
      of batchnorm vs. groupnorm.
    \item a study of the different $\alpha$-matte losses ($\mathcal{L}_1$, gradient,
      laplacian pyramid, compositing loss).
    \item a study of the potential benefit of also predicting $\fgt$ and $\bgt$
      alongside $\alpha$ and the possible losses associated with this ($\mathcal{L}_1$
      loss and exclusion loss).
\end{enumerate}

Related works are presented in section~\ref{sec:related-works}, our proposed method is proposed in section~\ref{sec:method}. The experimental studies and their results are discussed in section~\ref{sec:results}. Our resulting network achieves
the state of the art performance on the Adobe Com\-po\-si\-tion-1k dataset for
alpha matte and composite colour quality. It is also the current best performing
method on the \url{alphamatting.com} online evaluation.

\section{Related Works}\label{sec:related-works}

\paragraph{Alpha Matting.}

Cutting out objects has been a key task in image editing and visual media
post-production for some time now and the way this is currently done in the
industry is through the use of greenscreen and manual rotoscopy, which are the
only reliable methods available to date. Early attempts at automating this task
without a greenscreen include the development of sampling
strategies~\cite{sharedMatting,comprehensiveSamplingMatting,klDivergenceMatting}
to find suitable candidate pairs for the foreground and background colours $\fgt_i$
and $\bgt_i$ at each pixel $i$. These samplings techniques typically lack spatial
consistency and it is only in the 2007 landmark paper by Levin
\etal~\cite{ClosedFormMattingPAMI}, that a comprehensive spatial prior model for
Matting was first proposed. This smoothness model has since then been widely
adopted in the literature~\cite{He11,Shahrian13,karacan2017alpha} and it is only
with the advent of deep convolutional neural networks that more expressive
spatial priors have been designed~\cite{DeepImageMatting,IndexMatting}.

\paragraph{Deep Alpha Matting Network Architectures.}

The $\alpha$-matting networks have mainly been designed as adaptions of
segmentation architectures. In 2017 by Xu \etal \cite{DeepImageMatting} first
proposed in \textit{Deep Image Matting} to adapt the FCN architecture of
~\cite{FCN} and adjoin, to the RGB image, the user trimap as an
extra channel. From there on, the focus has shifted towards improving the core
encoder-decoder architecture, so as to increase the resolution of the output
predictions. For instance, in \textit{VDRN Matting} (2019)~\cite{VDRNMatting}
deeper encoder and decoder residual networks have been proposed. In
\textit{IndexNet} (2019) \cite{IndexMatting}, the encoder-decoder uses a
learnable index pooling. In \textit{Context Aware Matting} (2019)
\cite{ContextMatting}, two encoders with atrous convolutions are fused
together. Some other works have also tried to better exploit the user
generated trimap. In \textit{GCA Matting} (2020) \cite{GCAMatting}, an attention
mechanism is designed to use the trimap as a guide. In ~\cite{DisentangledMatting}, a two stage strategy is proposed to first refine the original trimap then proceed with alpha matting.

\paragraph{Losses.}

The exact nature of the output $\alpha$-matte is a key aspect of matting. On one
hand, $\alpha = \{0,1\}$ represents a binary label field similar to the one found
in binary segmentation. On other hand, the $\alpha$-matte is also a continuous
field with values between 0 and 1, which shares some resemblance to natural
images. This opens up a wide range of possible losses. In \textit{Deep Image
  Matting}, Xu \etal\cite{DeepImageMatting} originally proposed a simple Huber
loss on $\alpha$. In later works~\cite{DisentangledMatting,ContextMatting,GCAMatting} the $\mathcal{L}_1$ loss was
preferred instead. The absolute values of $\alpha$ may however not be as
important as the gradient of $\alpha$. Indeed, errors in the reproduction of the
hair strands shapes are more noticeable than slight errors in the overall
opacity level. Gradient fidelity is in fact one of the commonly used quality
metric in image matting benchmarks. Gradient related losses include the use of
$\mathcal{L}_1$ gradient loss~\cite{LearningBasedSamplingMatting} and the use of the pyramid Laplacian loss (see Context
Aware Matting \cite{ContextMatting}). Beyond these set losses, Lutz \etal
(2018)~\cite{AlphaGAN} have also proposed a discriminative loss on $\alpha$ as
part of their GAN architecture.

The matting problem is however not limited to the $\alpha$ field alone as
$(\alpha,F,B)$ are interdependent. In \textit{Deep Image Matting}, Xu
\etal\cite{DeepImageMatting} thus proposed to combine their Huber loss on
$\alpha$ with a \textit{compositing} loss $\mathcal{L}_c$, which is defined as the
reconstruction error $\mathcal{L}_c(\alphap) = \sum_i \| \cgt_i - \left(\alphap_i \fgt_i +
(1-\alphap_i) \bgt_i \right)\|_1$ when using the ground truth foreground and
background colours $\fgt_i$ and $\bgt_i$ at pixel $i$. This loss has then
been used in later works~\cite{AlphaGAN,LearningBasedSamplingMatting,VDRNMatting}. 
\paragraph{Foreground \& Background Predictions.}

A few prior non-deep learning methods have proposed to predict $\fgt$ and
$\bgt$ along side $\alpha$~\cite{SimultaneousMatting,sharedMatting,IfmMatting} but it
is only very recently that Hou and Liu~(2019)~\cite{ContextMatting} and
Tang~\etal~(2019)~\cite{LearningBasedSamplingMatting} have started to look into
jointly estimating fg/bg colours with alpha in CNNs. In \cite{ContextMatting},
 $\fgt$ and $\alpha$ are decoded in sequence from the same shared decoder. In
\cite{LearningBasedSamplingMatting}, the estimation is also done in the sequence
$\bgt$, $\fgt$ and $\alpha$, using this time three full separate encoder-decoder
networks. The main drawback here is that sequential estimation
depends on the success of each of the individual predictions. Also, stacking
full encoder-decoders as in~\cite{LearningBasedSamplingMatting} results in a
very deep and large networks.

These works have chosen the $\mathcal{L}_1$ loss on the predicted $\fgt$ and $\bgt$.  In
\cite{ContextMatting}, they also use a VGG16 Features Loss on the predicted
pre-multiplied Foreground $\alphap \fp$.

Also predicting $\bp$ allows \cite{LearningBasedSamplingMatting} to introduce a new composition loss for the predicted $\fp$ and $\bp$, based on the
ground-truth values of $\alphagt$: $\mathcal{L}_{c}(\fp,\bp) = \sum_i \|\cgt_i -  \left(\alphagt_i \fp_i +
(1-\alphagt_i) \bp_i\right) \|_1$.

Other losses are however possible. One particular loss we want to investigate is
the exclusion loss ($\mathcal{L}_{\mathrm{excl}}(\fp,\bp) = \sum_i \|  \nabla \fp_i \|_1  \| \nabla \bp_i \|_1$), which has been introduced in a similar form in the reflection removal literature~\cite{perceptualReflectionRemoval}
to enforce a clean separation between $\fgt$ and $\bgt$ and avoid that structures of
the original image to leak into both $\fgt$ and $\bgt$.

\section{Proposed Approach}\label{sec:method}

\subsection{Network Architecture}

Similarly to most previous works, we use an encoder-decoder with
Unet~\cite{unet} style architecture. The main difference is that our network
also predicts $\fgt$ and $\bgt$ directly from this single encoder-decoder. Jointly
estimating for $\alpha$, $\fgt$ and $\bgt$ is motivated by the applications in
compositing requiring an estimate for the foreground, and also by results in
Multi-Task learning~\cite{ruder2017overview}. We choose to do this in the simplest way by
extending the number of output channels from one to seven (1 for $\alpha$, 3 for
$\fgt$ and 3 for $\bgt$). In contrast to previous works~\cite{ContextMatting} and
\cite{LearningBasedSamplingMatting}, which adopt a sequential prediction, our
approach requires minimal extra parameters and avoids the delicate chaining of
estimations.

\paragraph{The encoder architecture} is ResNet-50, and the weights are
initialised by pre-training on ImageNet~\cite{imagenet} for
classification~\cite{weightstandardization}. Two modifications are made to the
encoder network.

First, we increase the number of input channels from 3 to 9 to allow for the
extra trimap.  We encode the trimap using Gaussian blurs of the definite
foreground and background masks at three different scales (in a similar way to
the method of~\cite{LeInteractiveSelection} in interactive segmentation). This
encoding differs from existing approaches in deep image matting, as they usually
encode the trimap as a single channel with value 1 if foreground, 0.5 for
unknown and 0 for background.

Second, we remove the striding from `layer 3' and `layer 4' of ResNet-50 and
increase the dilation to 2 and 4 respectively, in a similar way to what was
proposed in AlphaGAN. In this way, the information can be processed at the
highest scales without lowering the tensor resolution.

\paragraph{The Decoder} then passes encoder features to a Pyramid Pooling
layer~\cite{PSPNet}. The pooled features are fed into a decoder which contains
seven convolutional layers interleaved with three bilinear upsampling layers and
skip connections. We provide full network details in the supplementary materials.

\paragraph{The Output Layer} contains 7 channels for $\alpha, \fgt, \bgt$. 
Experiments in section~\ref{sec:results} show that clamping the values of $\alpha$
between 0 and 1 with a hardtanh activation as in Deep Image Matting~\cite{DeepImageMatting} gives a small improvement over using
a sigmoid function as in other previous works~\cite{VDRNMatting,GCAMatting,LearningBasedSamplingMatting,AlphaGAN,ContextMatting,DisentangledMatting}. The $\fgt,\bgt$ channels also go through
a sigmoid activation functions so a to also stay in the $[0,1]$ range.

\subsection{Batch Normalisation vs. Group Normalisation}

Training Matting networks can take a long time and small implementation details
can sometimes matter. One such detail that we discovered to have critical
importance is the mini-batch size. All prior matting works have adopted a
relatively small mini-batch size of around 6-16~\cite{ContextMatting,GCAMatting}. But through observing opensource re-implementations\cite{pytorchDIM} of the \textit{Deep Image Matting} method we found that a mini-batch size of one can greatly increase the network
accuracy in the original \textit{Deep Image Matting} \cite{DeepImageMatting}
paper.

A mini-batch size of one is however incompatible with our ResNet-50 encoder as
ResNet-50 uses Batch-Normalisation, which assumes i.i.d batches larger than size
1. We propose thus to use instead Group Normalisation (32 channels per group)
with Weight Standardisation~\cite{groupnorm,weightstandardization}.

\subsection{$\fgt,\bgt,\alpha$ Losses} 

As discussed in Section~\ref{sec:related-works}, numerous losses have been proposed to improve the training of alpha matting networks. To train our model we use a sum
of all of these losses; the $\mathcal{L}_1$ loss on alpha $\mathcal{L}^{\alpha}_1$, the composition loss $\mathcal{L}^{\alpha}_c$, the gradient
loss $\mathcal{L}^{\alpha}_g$, and the Laplacian pyramid loss that is computed over 
multiple scales $s$ of the Laplacian pyramid $L^s_{\mathrm{pyr}}$. For training the foreground and background we also use a sum of loss functions; $\mathcal{L}^{\mathrm{FB}}_1$, laplacian loss $\mathcal{L}^{\mathrm{FB}}_{\mathrm{lap}}$, compositional loss $\mathcal{L}^{\mathrm{FB}}_c$, and a gradient exclusion loss $\mathcal{L}^{\mathrm{FB}}_{\mathrm{excl}}$. See Table~\ref{tab:losses} for the definition of each loss function. 
Our final loss function is 
\begin{equation}
    \mathcal{L}_{FB\alpha} = \mathcal{L}^{\alpha}_1 +\mathcal{L}^{\alpha}_c+\mathcal{L}^{\alpha}_g+\mathcal{L}^{\alpha}_{\mathrm{lap}} + 0.25\left(
    \mathcal{L}^{\mathrm{FB}}_1 + 
    \mathcal{L}^{\mathrm{FB}}_{\mathrm{lap}} +
    \mathcal{L}^{\mathrm{FB}}_{\mathrm{excl}} +
    \mathcal{L}^{\mathrm{FB}}_c
    \right)
\end{equation}

\begin{table}[t]
\centering
  \caption{Training Loss Functions. }
\label{tab:losses}
\begin{tabular}{ll}
\toprule
$\alphagt$ Losses  & $\fgt ,\bgt$ Losses         \\
\midrule
\begin{minipage}{.5\linewidth}
  $\begin{aligned}
    \mathcal{L}^{\alpha}_1 &= \sum_i \left\| \alphap_i - \alphagt_i \right\|_1   \\
    \mathcal{L}^{\alpha}_c &= \sum_i \left\| \cgt_i -  \alphap_i \fgt_i - (1-\alphap_i) \bgt_i \right\|_1  \\
    \mathcal{L}^{\alpha}_{\mathrm{lap}} &=\sum_{s=1}^5 2^{s-1} \left\| L_{\mathrm{pyr}}^s(\alphagt) - L_{\mathrm{pyr}}^s(\alphap) \right\|_1 \\
    \mathcal{L}^{\alpha}_g &= \sum_i \left\| \nabla \alphap_i - \nabla \alphagt_i \right\|_{1} \\
\end{aligned}$
\end{minipage}
&
\begin{minipage}{.5\linewidth}
  $\begin{aligned}
    \mathcal{L}^{\mathrm{FB}}_1 &= \sum_i \left\| \fp_i - \fgt_i \right\|_1 + \left\| \bp_i - \bgt_i \right\|_1  \\
    \mathcal{L}^{\mathrm{FB}}_{\mathrm{excl}} &= \sum_i \left\| \nabla \fgt_i \right\|_1 \left\| \nabla \bgt_i \right\|_1 \\
\mathcal{L}^{\mathrm{FB}}_c &= \sum_i \left\| \cgt_i -  \alphagt_i \fp -  (1-\alphagt_i) \bp ) \right\|_1 \\
\mathcal{L}^{\mathrm{FB}}_{\mathrm{lap}} &= \mathcal{L}^{\fgt}_{\mathrm{lap}} + \mathcal{L}^{\bgt}_{\mathrm{lap}}  \\
 \end{aligned}$ 
\end{minipage} \\
 \bottomrule
\end{tabular}
\end{table}

\subsection{$\fp,\bp,\alphap$ Fusion}

One limitation of our joint prediction approach is that our predictions for
$\alphap$, $\fp$ and $\bp$ are decoupled and, even if they are based on the same
decoder, the relationship given by the compositing Equation~\ref{eq:matting} is
not explicitly enforced. We propose here a fusion mechanism based on the maximum
likelihood estimate of $p(\alphagt, \fgt, \bgt | \alphap, \fp, \bp)$. By
assuming independence of the prediction errors and ignoring any spatial
dependence between pixels, we can build a simplified likelihood model, derived
from the individual and reconstruction losses:
\begin{equation}
p(\alphagt, \fgt, \bgt | \alphap, \fp, \bp) \propto p(\alphagt| \alphap)p(\fgt |
\fp)p(\bgt | \bp)p(\alphagt,\fgt, \bgt)
\end{equation}
Assuming Gaussian distributions for the predictions and reconstruction errors:
{ \small
\begin{align*}
 p(\fgt | \fp) &\propto \exp\left(-\frac{\|\fgt - \fp \|_2^2}{2\sigma_{FB}^2} \right)  &
                                                                                         p(\bgt | \bp) &\propto\exp\left( -\frac{\|\bgt - \bp \|_2^2}{2\sigma_{FB}^2}
                                                                                                           \right) \\
 p(\alphagt | \alphap) &\propto \exp\left( -\frac{(\alphagt - \alphap )^2}{2\sigma_\alpha^2} \right)  &
 p(\alphagt , \fgt, \bgt) &\propto \exp\left( -\frac{\|\cgt -  \alphagt\fgt - (1-\alphagt)\bgt \|_2^2}{2\sigma_{C}^2}\right) \\\end{align*}}
Essentially, we have simplified the model by ignoring the spatial losses
(gradient and Laplacian pyramid) and replaced the $\mathcal{L}_1$ losses by $\mathcal{L}_2$ losses.
This simplified model still yields a non-linear optimisation because of the
reconstruction term is non-linear but we can adopt an iterative block solver
approach.  Starting at $\fp^{(0)}=\fp,\bp^{(0)}=\bp,\alphap^{(0)}=\alphap$, the
update equations are as follows:
\begin{align}
  \fp^{(n+1)} &= \fp +
  \frac{\sigma_F^2}{\sigma_\cgt^2} \alphap^{(n)} \left( \cgt - \alphap^{(n)} \fp^{(n)} - (1-\alphap^{(n)})\bp^{(n)} \right) \\
  \bp^{(n+1)} &= \bp + \frac{\sigma_B^2}{\sigma_\cgt^2}
  (1-\alphap^{(n)})\left(\cgt -  \alphap^{(n)} \fp^{(n)} - (1-\alphap^{(n)})\bp^{(n)} \right) \\
  \alphap^{(n+1)} &= \frac{\alphap^{(n)} + \frac{\sigma_\alpha^2}{\sigma_\cgt^2}
    (\cgt-\bp^{(n+1)})^{\top}(\fp^{(n+1)}-\bp^{(n+1)}) }{1 + \frac{\sigma_\alpha^2}{\sigma_\cgt^2}
    (\fp^{(n+1)}-\bp^{(n+1)})^{\top}(\fp^{(n+1)}-\bp^{(n+1)}) } 
\end{align}
These equations are related to the Bayesian Matting estimation scheme~\cite{bayesianMatting},
except that the covariance matrices for $F,B,C$ are not available. In practice,
we found that 1 iteration through these block updates was enough. These
equations give us a simple mechanism to fuse the 3 predictions by taking into
account the matting model of Eq.~\ref{eq:matting}. Our experiments in section~\ref{sec:results}, with $\sigma^2_{C},\sigma^2_{F},\sigma^2_{B}=1,\sigma^2_{\alpha}=10$, show
that these updated estimates reliably produce better results.

\subsection{Training Dataset and Modifications}

The large Deep Image Matting dataset constructed by Xu et
al.~\cite{DeepImageMatting} has been instrumental for training state of the art
matting algorithms in recent years. The dataset is a collection of 431
foreground and alpha channel pairs. Training samples are created by using the
alpha channel and the compositing equation~\ref{eq:matting} to composite the foreground
onto a randomly chosen background from the MSCOCO dataset~\cite{mscoco}. %

\begin{figure}[t]
 \centering
\setlength{\tabcolsep}{0.1em}
\begin{tabular}{ccccc}
  \includegraphics[width=.2\linewidth]{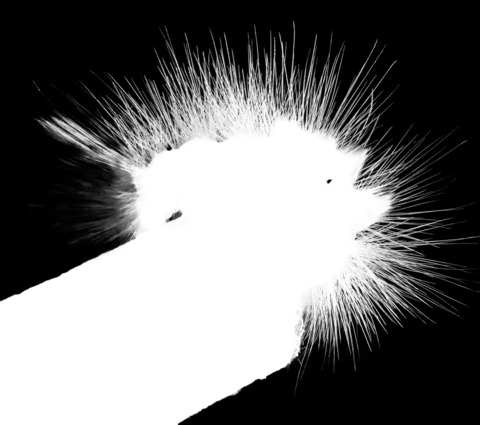}  &
  \includegraphics[width=.2\linewidth]{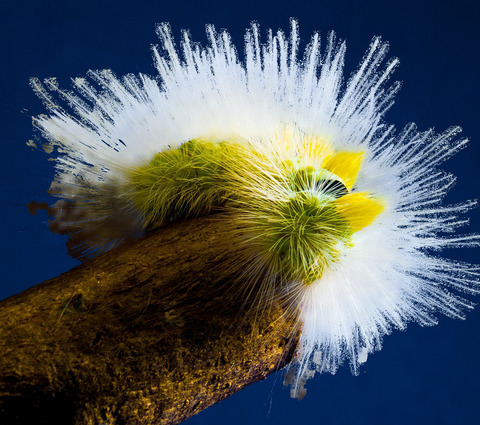}  &
    \includegraphics[width=.2\linewidth]{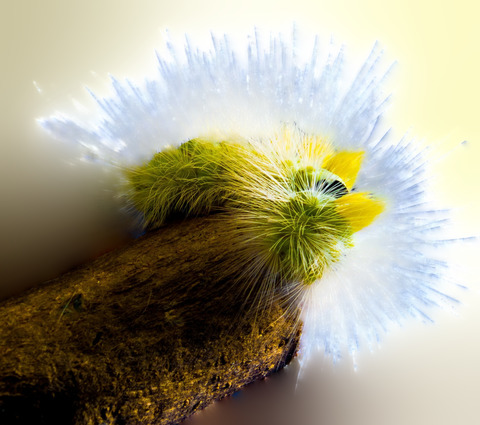}
  &
    \includegraphics[width=.2\linewidth]{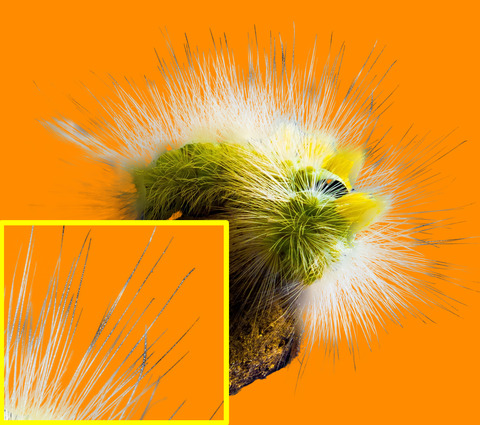}
  &
    \includegraphics[width=.2\linewidth]{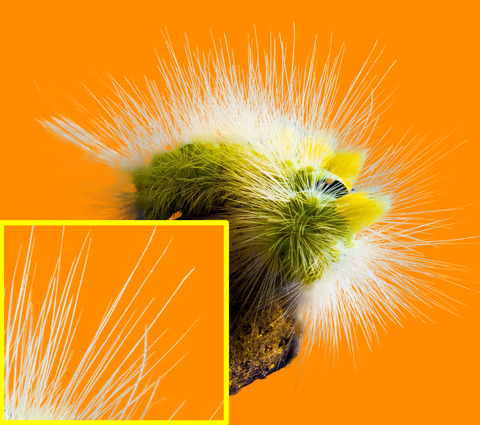} \\
\end{tabular}
\caption{\textbf{Colour-spill from Data Augmentation.} From left to right: the
  ground-truth $\alpha$-matte and ground-truth Fg colours as provided
  in~\cite{DeepImageMatting}, our re-estimated extended Fg colours, a composite
  image of the provided $\fgt$ after a resize augmentation on an orange background,
  our composite after the same resize operation. Resizing can cause the undefined
  Fg colours (in dark blue) to spill into the composite
  image.}\label{fig:fg-dataset-resize}
\end{figure}

The issue is that the foreground images provided are only valid for non-zero
alpha values and data augmentation during training can `spill' invalid colours
into these areas. For example, operations like resize and rotation re-sample
pixels indiscriminately from both valid and invalid regions (see
Figure~\ref{fig:fg-dataset-resize}). To remedy this, we re-estimated the foreground colours
for all images, using Levin's $\fgt,\bgt$ estimation
technique~\cite{ClosedFormMattingPAMI}. This allowed us to extend the foreground
estimation to the entire picture, and not only areas where $\alpha>0$. That way,
augmentation techniques become possible and makes this modified Deep Image
Matting training set suitable for foreground prediction.

\subsection{Training Details}

Similarly to what can be found in the literature, our training patches of
dimensions $640\times 640, 480\times 480, 320 \times 320$ are randomly cropped
from unknown regions of the trimap. The training trimaps are generated from the
ground truth $\alpha$-mattes by random erosion and dilation of 3 to 25
pixels. For data augmentation, we adopt random flip, mirroring, gamma, and brightness augmentations. To further increase the dataset
diversity, we randomly composite a new foreground object with 50\% probability, as in ~\cite{LearningBasedSamplingMatting}. The training data is 
shuffled after each epoch. Additionally, every second mini-batch is 
sampled from the $2\times$ image of the previous image, so as to increase the
invariance to scale.

We use step-decay learning rate policy with the RAdam optimiser~\cite{RAdam},
with momentum and weight decay set to 0.9 and 0.0001 respectively. The initial
learning rate is set at $10^{-5}$ and then dropped to $10^{-6}$ at 40 epochs and
fine-tuned for 5 more epochs. We apply weight decay of $0.005, 10^{-5}$ to
convolutional weights, and the GN parameters respectively. The training process
takes 16 days with a single 1080ti GPU. During inference, the full-resolution
input images and trimaps are concatenated as  4-channel input
and fed into the network.

\subsection{Test Time Augmentation (TTA)}

Convolutional neural networks are not invariant to flipping, rotation, zooming
and cropping of the input. Randomly augmenting the training samples in these
non-destructive ways has the effect of enlarging the training set and improve the final network
accuracy~\cite{LearningBasedSamplingMatting,GCAMatting}. However, training with
augmented examples does not ensure total invariance to the
augmentations. Predictions from augmented inputs are normally distributed, and
the average of the transformations tends to the true
value~\cite{aleatoricTTA}. Test-time eight-way rotation of the input image and
trimap was used by Tang et al.~\cite{VDRNMatting} but its influence on alpha
matte accuracy was not published. We use a comprehensive test-time augmentation, combining rotation, flipping and scaling,  and results are shown in Tables \ref{tab:comp1k_alpha_compare} and \ref{tab:comp1k_fg_ablation}.

\section{Experimental Results}
\label{sec:results}

In this section we report quantitative and qualitative results of our model. We perform an ablation study of our model and we also show state of the art results in comparisons to existing matting methods. 
We use the Composition-1k~\cite{DeepImageMatting} dataset for testing, as it contains 1000 testing images and other methods have reported their results on it for comparison. The dataset provides 50 ground truth foreground images and alpha mattes, and they are composed onto 20 different backgrounds each from Pascal~\cite{pascal2012}. 

We evaluate the results of both the alpha matte and also the combined $\alphap\fp$ foreground composite; supplemental results of the background prediction can be found in the supplemental material. 

The alpha matte results are computed using four standard metrics~\cite{alphamattingcom}, Sum of Absolute Differences (SAD), Mean Squared Error (MSE), Gradient Error (GRAD) and Connectivity Error (CONN). The gradient and connectivity metrics were shown at the time to be more aligned to human perception of matte quality. 
To measure foreground composite results we measure the MSE and SAD of the predicted $\alphap\fp$ and the ground truth $\alphagt\fgt$.

Most of the ablation study is done over a 20 epochs training. The training was pushed to 45 epochs, including 5 epochs of fine-tuning at a $10^{-6}$ learning rate, on the most complete models. 

Because of the scale of the experiments ---each training taking two to four weeks to complete--- the experiments have only been done on a single training run. Precise confidence intervals are thus not known but our earlier experiments seem to suggest that most of the observations made below are consistent.

\subsection{Evaluating the $\alpha$ Loss Functions}
As discussed previously, four existing alpha matting losses emerge as reasonable options to be summed for training our network. These are the $\mathcal{L}_1$ loss, the composition loss, the laplacian loss and the gradient loss. Here we evaluate our network trained with combinations of these losses, models (1-4) in Table~\ref{tab:comp1k_alpha_ablation}. We see that the compositing loss decreases errors for all metrics except for the gradient error. The laplacian loss proposed by Hou \etal~\cite{ContextMatting} gives a significant reduction in errors across all metrics. We note this network, training and loss configuration is enough to achieve state of the art results, see Table~\ref{tab:comp1k_alpha_compare}. The gradient loss proposed by Tang et al.~\cite{LearningBasedSamplingMatting} seems to increase the errors on all metrics. This was also reported in~\cite{GCAMatting}. As this discovery was made late in our research, the gradient loss is included in all of our subsequent models. We leave the further examination of this loss to future work. 

We also perform an experiment on the choice of alpha channel activation function (Table~\ref{tab:comp1k_alpha_ablation}). A clipping activation was used in the original Deep Image Matting work~\cite{DeepImageMatting}, yet all subsequent works used a sigmoid activation, without reference to this change. However, we find that the sigmoid activation underperforms compared to the clipping activation (models (6) vs. (7)), see also Table~\ref{tab:comp1k_fg_ablation} model (9) vs. (10). This also moves against the trend to use sigmoid for other image-to-image translation tasks~\cite{pix2pix,gatedinpainting}, however in these cases the rgb values are not usually 0 or 1 unlike alpha mattes, which are mostly 0 or 1. The sigmoid activation only achieves 0 and 1 for infinite valued inputs. Thus we use clipping activation for subsequent models.

\begin{table}[t]
\centering
\caption{Ablation study of $\alpha$-mattes results on the Composition-1k dataset.} 
\label{tab:comp1k_alpha_ablation}
\begin{tabular}{ccclcccc}
  \toprule
Model & Norm. & Batch-Size & Loss          & MSE & SAD & GRAD & CONN \\ 
\midrule
    \multicolumn{8}{l}{\textit{Training at 20 epochs:} \vspace{.3em}} \\
  (1) & BN            & 6         & $\mathcal{L}_1^\alpha$            &  11.2   & 36.3     & 14.9    & 32.5 \\
(2) & BN            & 6         & $\mathcal{L}_1^\alpha + \mathcal{L}_c^{\alpha}$        & 9.1    &    34.5 &   15.0    & 31.3 \\
(3) & BN            & 6         & $\mathcal{L}_1^\alpha + \mathcal{L}_c^{\alpha} + \mathcal{L}^{\alpha}_{\mathrm{lap}}$     & 7.4    & 33.5    & 12.9 & 28.5    \\
(4) & BN            & 6         & $\mathcal{L}_1^\alpha + \mathcal{L}_c^{\alpha} +
                                  \mathcal{L}^{\alpha}_{\mathrm{lap}} + \mathcal{L}^{\alpha}_{g}$ & 8.1    & 36.3     &    13.8  & 32.0 \\
(5) & GN            & 6         & $\mathcal{L}_1^\alpha + \mathcal{L}_c^{\alpha} +
                                  \mathcal{L}^{\alpha}_{\mathrm{lap}} + \mathcal{L}^{\alpha}_{g}$ & 10.3    & 36.2 & 15.1     &  32.0 \\

(6) & GN            & 1         & $\mathcal{L}_1^\alpha + \mathcal{L}_c^{\alpha} +
                                  \mathcal{L}^{\alpha}_{\mathrm{lap}} + \mathcal{L}^{\alpha}_{g}$ & 7.2    & 32.8 &    13.3  & 28.6 \\
(7) & GN            & 1         & $\mathcal{L}_1^\alpha + \mathcal{L}_c^{\alpha} +
                                  \mathcal{L}^{\alpha}_{\mathrm{lap}} + \mathcal{L}^{\alpha}_{g}$+ $\text{clip}_\alpha$ & 6.9      &  31.2   & 12.9 & 27.1 \\
\midrule
    \multicolumn{8}{l}{\textit{Training at 45 epochs:} \vspace{.3em}} \\
  \textbf{$\text{Ours}_\alpha$} &GN            & 1         & $\mathcal{L}_1^\alpha + \mathcal{L}_c^{\alpha} +
                                  \mathcal{L}^{\alpha}_{\mathrm{lap}} + \mathcal{L}^{\alpha}_{g}$+ $\text{clip}_\alpha$ & 5.3      &  26.5   & 10.6 & 21.8 \\
\bottomrule
\end{tabular}
\end{table}

\subsection{Batch-Size and BatchNorm vs. GroupNorm}

As discussed in the Proposed Approach section, we discovered that a mini-batch size of 1 greatly increases network accuracy for $\alpha$-matting. We report the results of an experiment on this in Table~\ref{tab:comp1k_alpha_ablation}. We use a model trained with a batch-size 6 and BatchNorm, model (4), as a baseline. BatchNorm is however, by definition, incompatible with training with mini-batch sizes of 1, so we use Group Normalisation with Weight Standardisation (WS)~\cite{groupnorm,weightstandardization} for single image mini-batches, model (6).

We also train an intermediate model (5) with GroupNorm and WS to isolate the effects of batch-size. As expected, we see a significant reduction in error from models (4,5) to model (6) showing that a batch-size of 1 is best suited to $\alpha$-matting. When comparing BN to GN with batch-size 6 it is clear that GroupNormalisation gives no hidden advantage to our hypothesis.

\subsection{Evaluating the Impact of Jointly Estimating $\fgt,\bgt,\alpha$}

In Table~\ref{tab:comp1k_fg_ablation} we examine the potential benefits of jointly estimating $\fgt,\bgt,\alpha$ over $\alpha$ alone. We measure the MSE and the SAD of the $\alpha \fgt$ composite and the SAD of the alpha alone, across the Composition-1k testing set. We also record an ablation study of our method to show the benefit of each component. We see that using our foreground and background loss $\mathcal{L}^{\mathrm{FB}}$ maintains alpha matte accuracy and allows for foreground prediction, model (6) vs. (9). Our foreground, background exclusion loss gives a minor benefit across all metrics, (8) vs. (9).

We observe an interesting trend in foreground prediction results from \textit{Context-Aware Matting}~\cite{ContextMatting}, see Figure~\ref{fig:fba1k}. The predicted foreground is quite poor in areas near the boundary or in highly transparent regions, and radically incorrect in nearby regions where $\alpha=0$. This causes issues of colour bleeding artifacts when compositing onto novel backgrounds. We believe this is due to them only computing the foreground loss on the regions where $\alpha>0$, and this leads to the network not prioritising foreground prediction in areas of very high transparency. We address this with our novel foreground dataset that has valid foreground ground truth for all pixels so we can compute the foreground prediction loss over the entire image. Our dataset also allows us to use the laplacian loss function for the foreground loss. In Table~\ref{tab:comp1k_fg_ablation} we see that training with the $\mathcal{L}^{\mathrm{FB}}$ throughout the entire image we improve the composite $\alpha F$ results, (10) vs. (11).

Our $\fgt, \bgt, \alpha$ fusion method, used at test-time, improves results for composite quality and for alpha matte quality, (11) vs. {$\textbf{Ours}_{\mathrm{FB}\alpha}$}. This small improvement has been consistently observed in our experiments.

\begin{table}[t]
\centering
\caption{Ablation study of foreground results on the Composition-1k dataset. Here $\mathcal{L}^{FB}= \mathcal{L}^{FB}_1 + \mathcal{L}_{\mathrm{lap}}^{FB} + \mathcal{L}_{\mathrm{c}}^{FB}$. In column two the * indicates that the $\mathcal{L}^{FB}_1, \mathcal{L}_{\mathrm{lap}}^{FB}$ are computed over the entire image as opposed to just the unknown region of the trimap. }
\label{tab:comp1k_fg_ablation}
\setlength{\tabcolsep}{.5em}
\begin{tabular}{clclcccc}
  \toprule
 Model & $+ \mathcal{L}_{FB}$ & $+ \mathcal{L}_{\mathrm{excl}}$
  & output &   \multicolumn{2}{c}{ $\alpha \fgt$}  &
                                                                         \multicolumn{2}{c}{ $\alpha$} \\
  & & &  &   SAD & MSE       & SAD & MSE  \\ 
  \midrule
   \multicolumn{4}{l}{ Closed-form Matting~\cite{ClosedFormMattingPAMI}} & 251.67 & 22.96 & 161.3 & 85.3 \\ 
   \multicolumn{4}{l}{ Context-Aware Matting~\cite{ContextMatting}} & 70.00 & 11.49 & 38.1 &  8.9\\ 

   \midrule
     \multicolumn{6}{l}{\textit{Training at 20 epochs:} \vspace{.3em}} &  &\\ 
  (6) &     N & N                                                   & sigmoid           & -  &- & 32.8& 7.2\\
  (8) &      Y&N                               & sigmoid           & 53.64        &  9.04    &  32.7& 9.0\\
  (9) &        Y&Y                    & sigmoid           &            52.87                   & 8.88  & 31.8 &  8.9 \\
  (7) &      N&N                                                    & clip              & -                                    & - & 31.2& 6.9\\
  (10) &         Y&Y                       & clip              &        50.69                  & 8.64   &    31.3  & 8.6 \\
  (11)  &         Y* &Y & clip              &                50.29                          &  8.48  & 32.1 &  8.5 \\ \midrule
   \multicolumn{6}{l}{\textit{Training at 45 epochs:} \vspace{.3em}} &  \\ 
  (11)&  Y*&Y  & clip      &  42.19      &     6.50  & 26.5 &  5.4 \\
   \textbf{$\text{Ours}_{\mathrm{FB}\alpha}$}& Y*&Y   & clip +fusion     & 39.21       &    6.19  & 26.4 & 5.4   \\
   \textbf{$\text{Ours}_{\mathrm{FB}\alpha}$} &Y* &Y   & clip +fusion +TTA  & 38.81        &   5.98  & 25.8 & 5.2\\
\bottomrule
\end{tabular}
\end{table}

\subsection{Comparison with Other Works}

We compare our models \textbf{$\text{Ours}_\alpha$} and $\text{Ours}_{\mathrm{FB}\alpha}$ with existing state of the art approaches to alpha matting.

\subsubsection{The Composition-1k dataset}

\begin{figure*}[p]
 \centering
\setlength{\tabcolsep}{0.1em}
\begin{tabular}{cccccc}
\includegraphics[width=.16\linewidth]{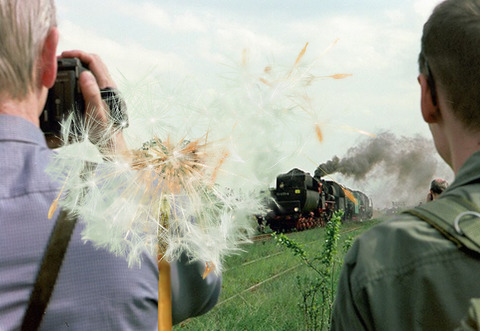}  &
\includegraphics[width=.16\linewidth]{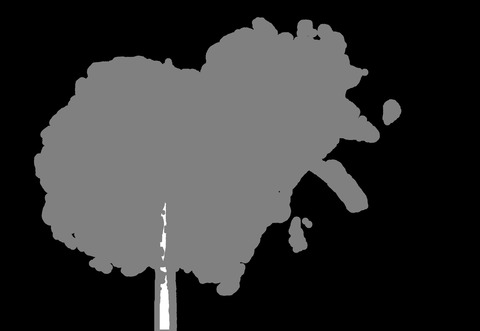}  &
\includegraphics[width=.16\linewidth]{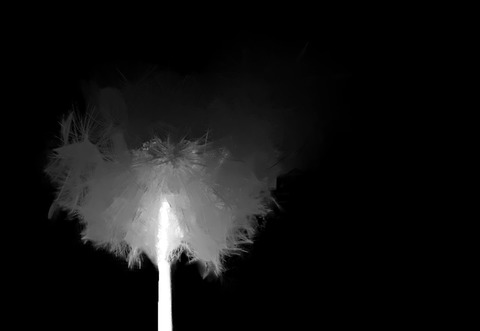}  &
\includegraphics[width=.16\linewidth]{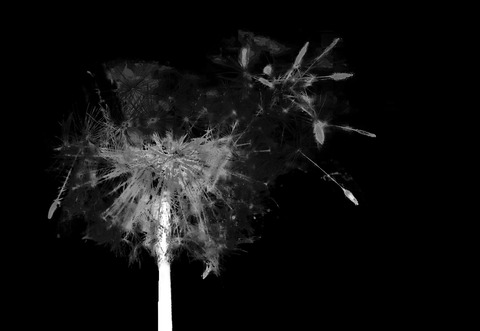}  &
\includegraphics[width=.16\linewidth]{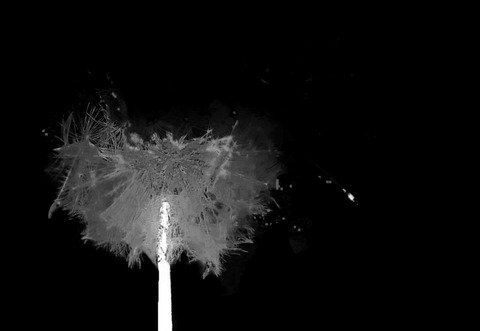}  &
\includegraphics[width=.16\linewidth]{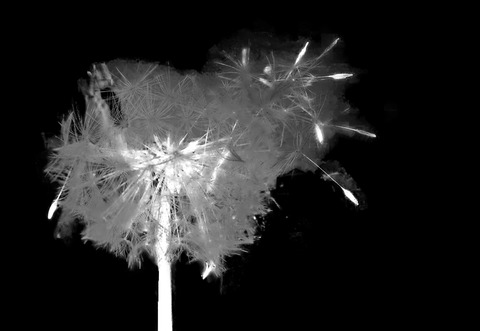}  \\
\tiny{Input Image} & \tiny{Trimap} & \tiny{Closed Form~\cite{ClosedFormMattingPAMI}} & \tiny{KNN~\cite{KnnMatting}} & \tiny{DCNN~\cite{DCNNMatting}} & \tiny{IFM~\cite{IfmMatting}} \\
\includegraphics[width=.16\linewidth]{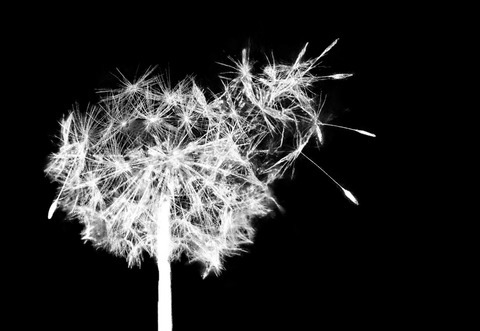}  &
\includegraphics[width=.16\linewidth]{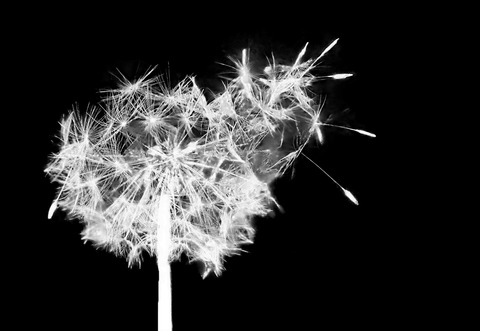}  &
\includegraphics[width=.16\linewidth]{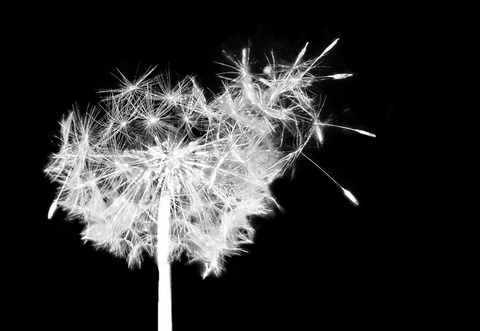}  &
\includegraphics[width=.16\linewidth]{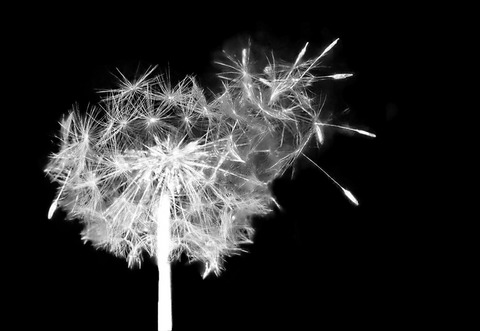}  &
\includegraphics[width=.16\linewidth]{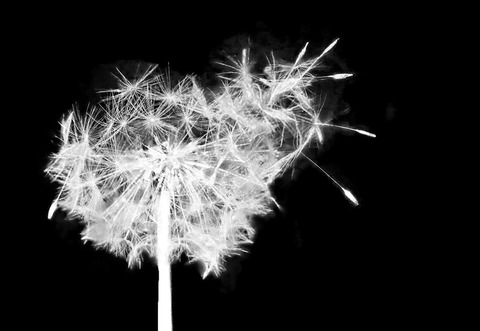} & 
\includegraphics[width=.16\linewidth]{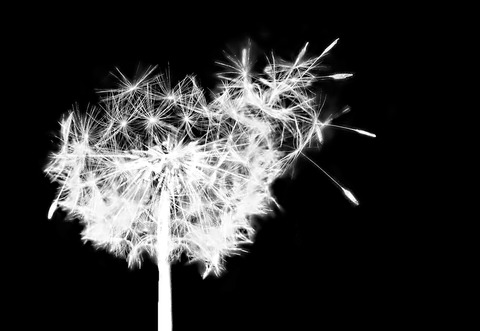}  \\
\tiny{Deep Matting~\cite{DeepImageMatting}} & \tiny{IndexNet~\cite{IndexMatting}} & \tiny{CA~\cite{ContextMatting} } & \tiny{GCA~\cite{GCAMatting}}& \tiny{$\text{Ours}_{\mathrm{FB}\alpha}$ TTA} & \tiny{Ground Truth} \\
\includegraphics[width=.16\linewidth]{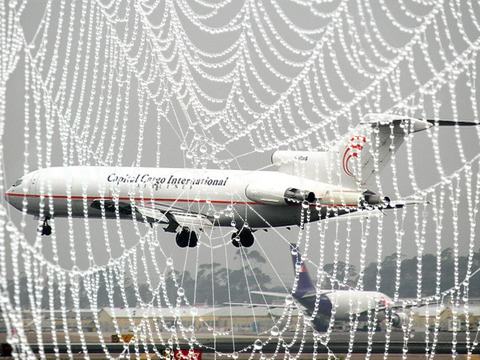}  &
\includegraphics[width=.16\linewidth]{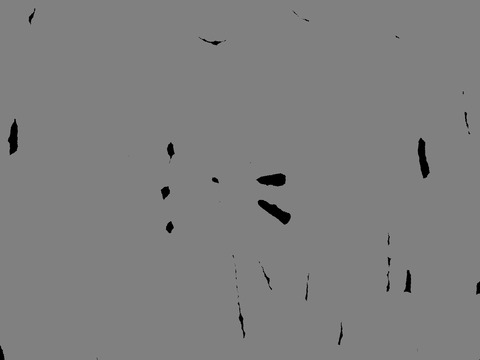}  &
\includegraphics[width=.16\linewidth]{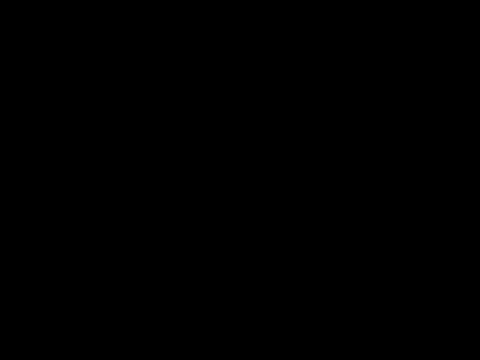}  &
\includegraphics[width=.16\linewidth]{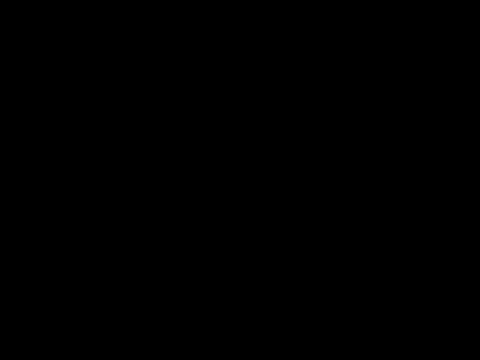}  &
\includegraphics[width=.16\linewidth]{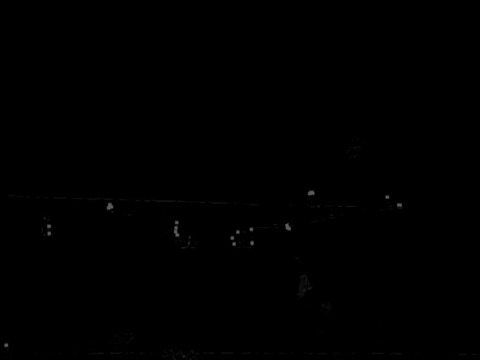}  &
\includegraphics[width=.16\linewidth]{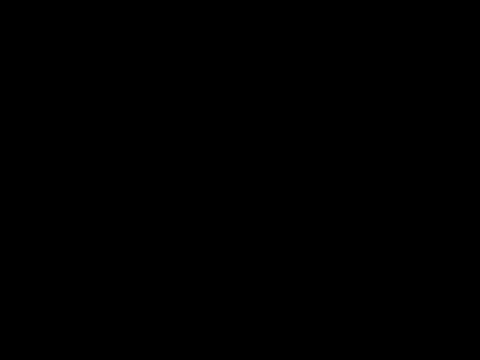}  \\
\tiny{Input Image} & \tiny{Trimap} & \tiny{Closed Form~\cite{ClosedFormMattingPAMI}} & \tiny{KNN~\cite{KnnMatting}} & \tiny{DCNN~\cite{DCNNMatting}} & \tiny{IFM~\cite{IfmMatting}} \\

\includegraphics[width=.16\linewidth]{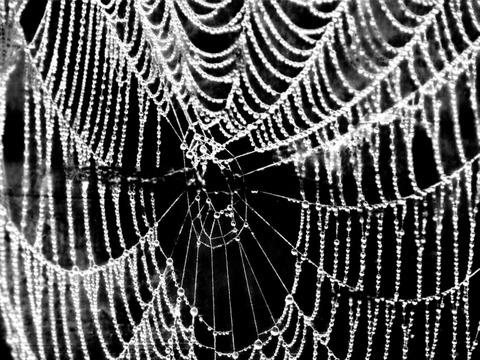}  &
\includegraphics[width=.16\linewidth]{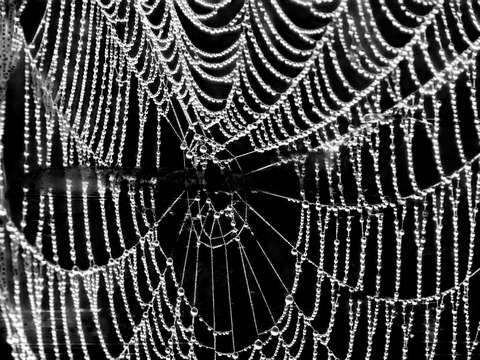}  &
\includegraphics[width=.16\linewidth]{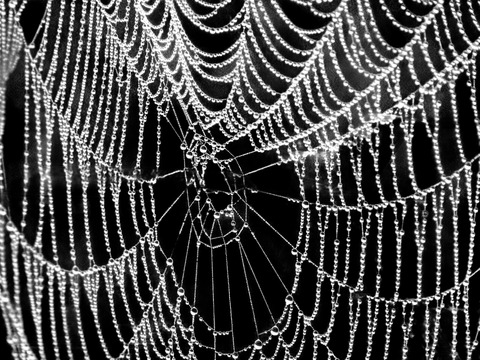}  &
\includegraphics[width=.16\linewidth]{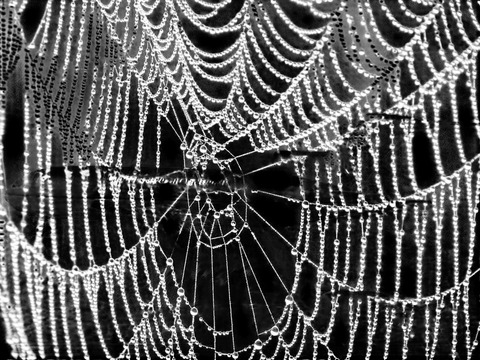}  &
\includegraphics[width=.16\linewidth]{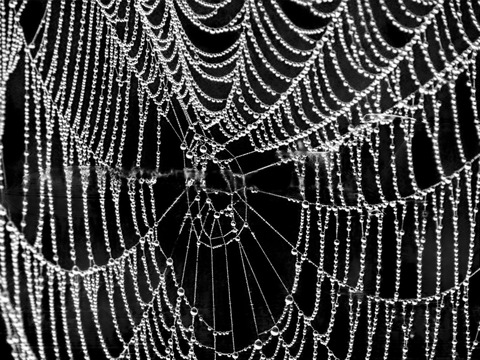} & \includegraphics[width=.16\linewidth]{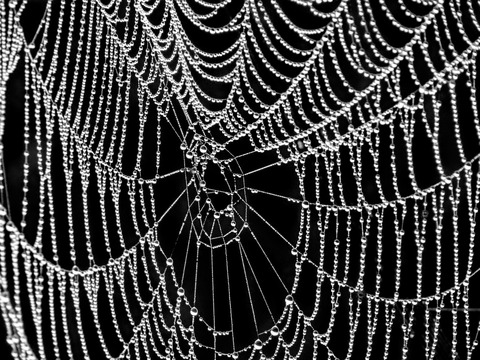}  \\

\tiny{Deep Matting~\cite{DeepImageMatting}} & \tiny{IndexNet~\cite{IndexMatting}} & \tiny{GCA~\cite{GCAMatting}} & \tiny{CA~\cite{ContextMatting}} & \tiny{$\text{Ours}_{\mathrm{FB}\alpha}$ TTA} & \tiny{Ground Truth}
\end{tabular}
\caption{Qualitative comparison of the alpha matte results on the Adobe Composition-1k test set~\cite{DeepImageMatting}.}\label{fig:alphas1k}
\end{figure*}

\begin{figure*}[p]
 \centering
\setlength{\tabcolsep}{0.1em}
\begin{tabular}{ccccc}
\includegraphics[height=.18\linewidth,angle=90]{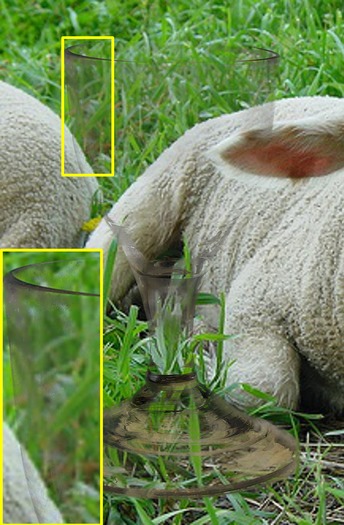}  &
\includegraphics[height=.18\linewidth,angle=90]{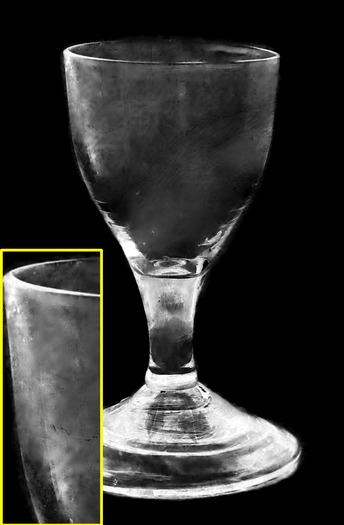}  &
\includegraphics[height=.18\linewidth,angle=90]{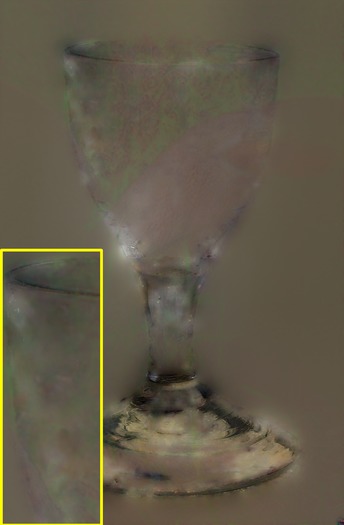}  &
\includegraphics[height=.18\linewidth,angle=90]{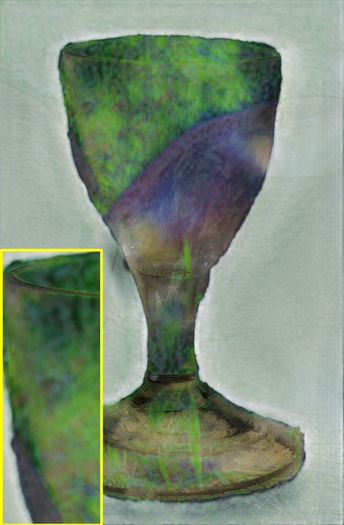}  &
\includegraphics[height=.18\linewidth,angle=90]{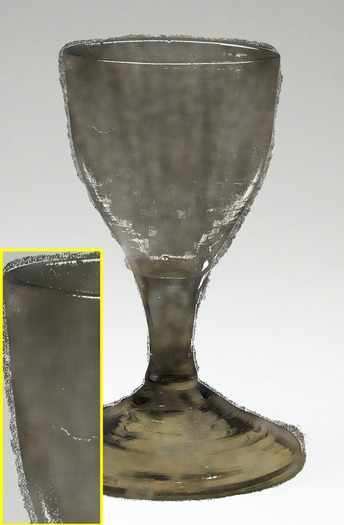}  \\
\tiny{Input Image} & \tiny{$\text{Ours}_{\mathrm{FB}\alpha}$ $\alphap$} & \tiny{$\text{Ours}_{\mathrm{FB}\alpha}$ $\fp$} & \tiny{CA~\cite{ContextMatting} $\fp$} & \tiny{Ground Truth $\fgt$} \\
\includegraphics[height=.18\linewidth,angle=90]{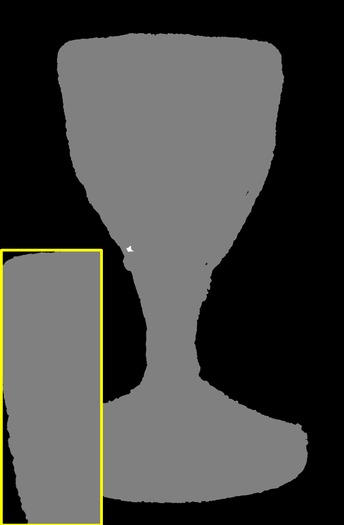}  &
\includegraphics[height=.18\linewidth,angle=90]{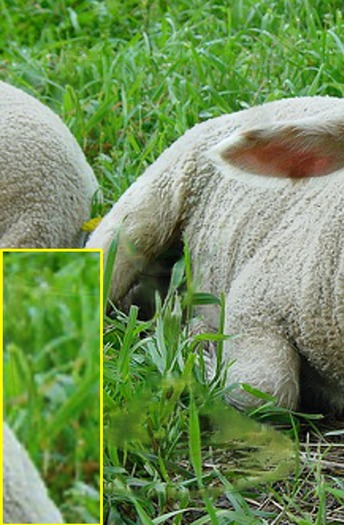}  &
\includegraphics[height=.18\linewidth,angle=90]{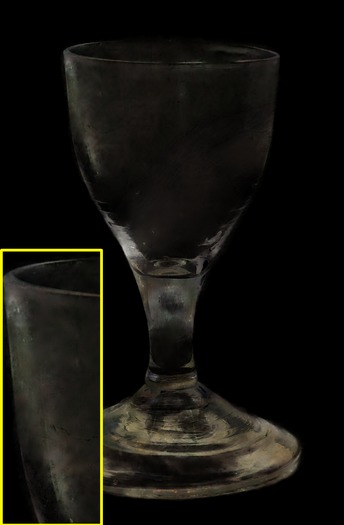}  &
\includegraphics[height=.18\linewidth,angle=90]{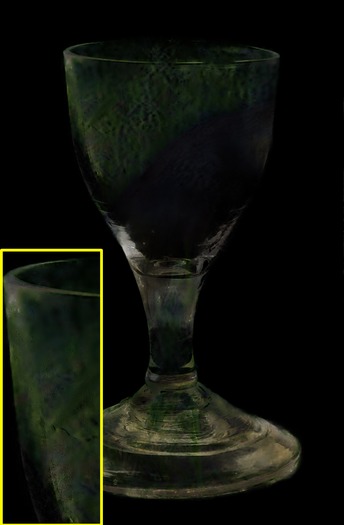}  &

\includegraphics[height=.18\linewidth,angle=90]{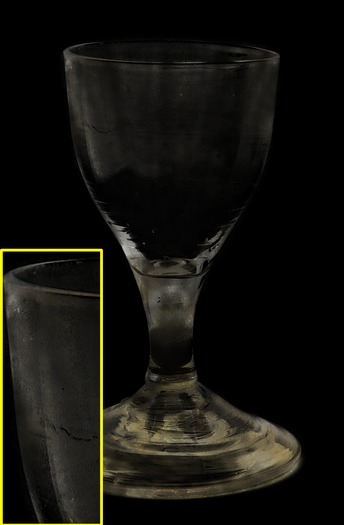}  \\
\tiny{Input Trimap} & \tiny{$\text{Ours}_{\mathrm{FB}\alpha}$ $\bp$} & \tiny{$\text{Ours}_{\mathrm{FB}\alpha}$  $\alphap \fp$} & \tiny{CA~\cite{ContextMatting} $\alphap \fp$} & \tiny{Ground Truth $\alphagt\fgt$} \\

\includegraphics[width=.18\linewidth]{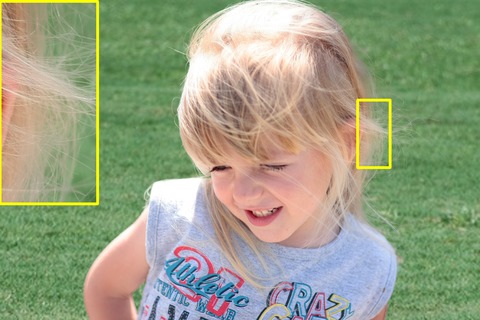}  &
\includegraphics[width=.18\linewidth]{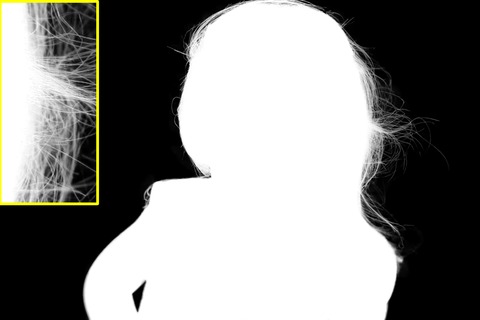}  &
\includegraphics[width=.18\linewidth]{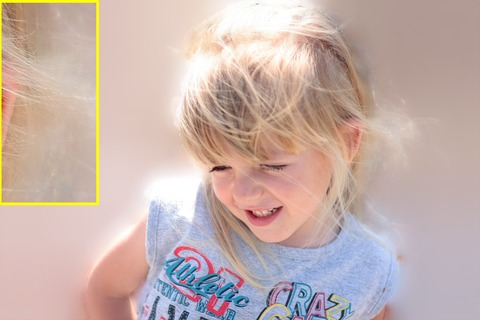}  &
\includegraphics[width=.18\linewidth]{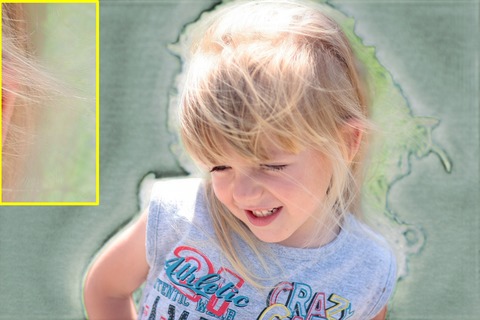}  &
\includegraphics[width=.18\linewidth]{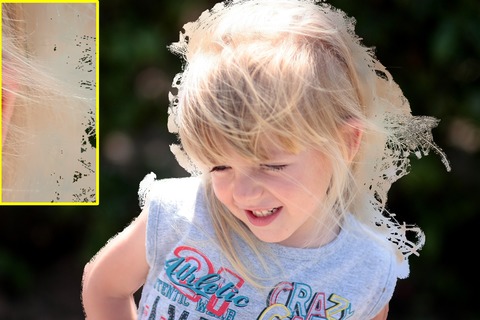}  \\
\tiny{Input Image} & \tiny{$\text{Ours}_{\mathrm{FB}\alpha}$ $\alphap$} & \tiny{$\text{Ours}_{\mathrm{FB}\alpha}$ $\fp$} & \tiny{CA~\cite{ContextMatting} $\fp$} & \tiny{Ground Truth $\fgt$} \\
\includegraphics[width=.18\linewidth]{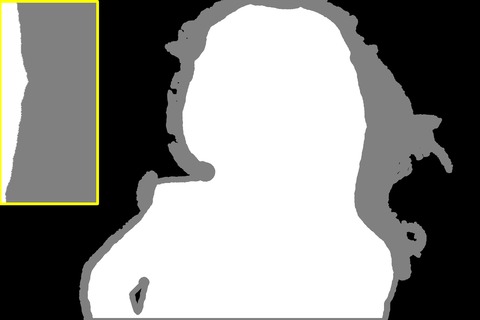}  &
\includegraphics[width=.18\linewidth]{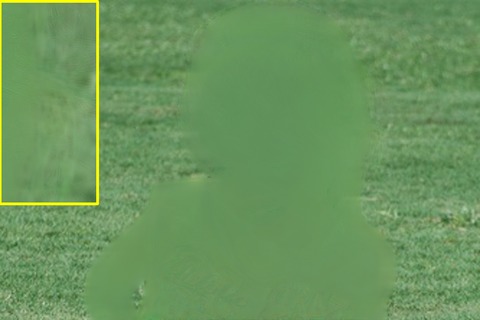}  &
\includegraphics[width=.18\linewidth]{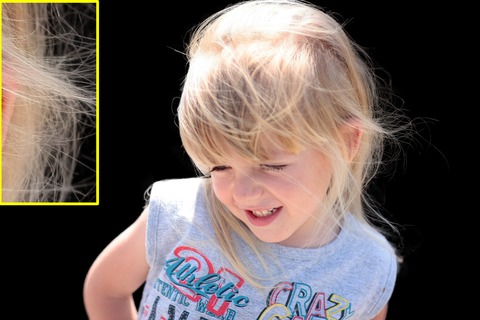}  &
\includegraphics[width=.18\linewidth]{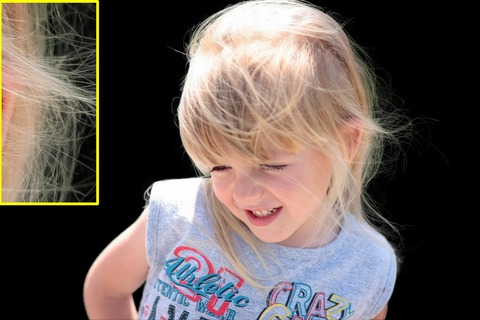}  &

\includegraphics[width=.18\linewidth]{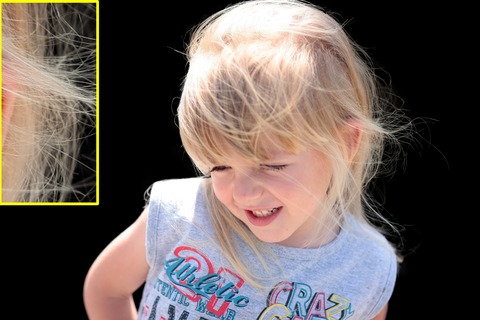}  \\
\tiny{Input Trimap} & \tiny{$\text{Ours}_{\mathrm{FB}\alpha}$ $\bp$} & \tiny{$\text{Ours}_{\mathrm{FB}\alpha}$  $\alphap \fp$} & \tiny{CA~\cite{ContextMatting} $\alphap \fp$} & \tiny{Ground Truth $\alphagt\fgt$} \\

\end{tabular}
\caption{Qualitative foreground, background and alpha matte results on the Adobe Composition-1k test set~\cite{DeepImageMatting}.}\label{fig:fba1k}
\end{figure*}

On the Composition-1k dataset, Table~\ref{tab:comp1k_alpha_compare}, each of our models significantly outperform previous approaches on all four metrics. We see that both our models \textbf{$\text{Ours}_{\alpha}$} and \textbf{$\text{Ours}_{\mathrm{FB}\alpha}$} achieve state of the art results. The model \textbf{$\text{Ours}_{\mathrm{FB}\alpha}$} which includes the foreground and background loss, with fusion, has improved connectivity and gradient measures but worse on the SAD measure.
Our best model \textbf{$\text{Ours}_{\mathrm{FB}\alpha}$ TTA} with test time augmentation achieves an average reduction in error of 34\% over the best performing method for each metric, and the gradient error has the largest relative improvement. 
We give a qualitative comparison of alpha matte quality in Figure~\ref{fig:alphas1k}. Our method can separate very fine structures from the background even where the foreground and background colours are very similar.  In Figure~\ref{fig:fba1k} we display the foreground, background and alpha outputs of our method and compare our predicted composite to that of Context-Aware matting~\cite{ContextMatting}. We see that for CA Matting the foreground prediction can be quite poor in areas where alpha is close to zero, and colour spill is still visible in the $\alphap\fp$ composite. Our method, on the other hand, matches the ground truth much more closely. In the first image our background prediction successfully removes the glass from the image, which shows our method has the additional use for removing semi-transparent objects from images, e.g. watermarks.

\begin{table}[t]
\centering
\caption{Alpha map results on the Composition-1k test set~\cite{DeepImageMatting}.}
\label{tab:comp1k_alpha_compare}
\begin{tabular}{lr<{\hspace{1em}}r<{\hspace{1em}}r<{\hspace{1em}}r}
\toprule
Method  &SAD  &  MSE {\scriptsize x$10^3$}  &  Gradient  & Connectivity  \\ \midrule
Closed-Form Matting~\cite{ClosedFormMattingPAMI}          & 168.1                    & 91.0                               & 126.9                         & 167.9                             \\ 
KNN-Matting~\cite{KnnMatting}                  & 175.4                    & 103.0                              & 124.1                         & 176.4                             \\ 
DCNN Matting~\cite{DCNNMatting}                 & 161.4                    & 87.0                               & 115.1                         & 161.9                             \\ 
Information-flow Matting~\cite{IfmMatting}     & 75.4                     & 66.0                               & 63.0                            & -                                 \\ 
Deep Image Matting~\cite{DeepImageMatting}           & 50.4                     & 14.0                               & 31.0                            & 50.8                              \\ 
AlphaGan-Best~\cite{AlphaGAN}                & 52.4                     & 30.0                               & 38.0                          & -                                 \\ 
IndexNet Matting~\cite{IndexMatting}             & 45.8                     & 13.0                               & 25.9                          & 43.7                              \\
VDRN Matting~\cite{VDRNMatting}                 & 45.3                     & 11.0                               & 30.0                            & 45.6                              \\ 
AdaMatting~\cite{DisentangledMatting}   & 41.7                    & 10.2                             &  {\color[HTML]{3531FF}16.9}                         & -                                 \\
Learning Based Sampling~\cite{LearningBasedSamplingMatting}      & 40.4                    & 9.9                              & -                             & -                                 \\ 

Context Aware Matting~\cite{ContextMatting}        & 35.8                     &  {\color[HTML]{3531FF}8.2}                            & 17.3                          & 33.2                              \\ 
GCA Matting~\cite{GCAMatting}                  &  {\color[HTML]{3531FF}35.3}                    & 9.1                              &  {\color[HTML]{3531FF}16.9}                         & {\color[HTML]{3531FF} 32.5}                             \\ 
\midrule
\textbf{$\text{Ours}_{\alpha}$} & 26.5  &  5.3 &   10.6 & 21.8\\ 
\textbf{$\text{Ours}_{\mathrm{FB}\alpha}$}  & 26.4  & 5.4  & 10.6 & 21.5    \\ 
\textbf{$\text{Ours}_{\mathrm{FB}\alpha}$ TTA}   & {\color[HTML]{FE0000}\textbf{ 25.8}} &  {\color[HTML]{FE0000}\textbf{ 5.2}} & {\color[HTML]{FE0000}\textbf{ 10.6}}& {\color[HTML]{FE0000}\textbf{ 20.8}}  \\ 
\bottomrule
\end{tabular}
\end{table}

\subsubsection{The alphamatting.com Dataset}

The \url{alphamatting.com} benchmark is an established online evaluation for natural image matting methods. It includes 27 training images and 8 testing images with 3 different kinds of trimaps, namely, “small”, “large” and “user”. Although there are only 8 testing images, it serves as an important tool for comparing alpha matting methods as the ground truth mattes are not released and the benchmark contains entries from most popular matting methods. In Table~\ref{tab:alphamattingcom} we see that our method, \textbf{$\text{Ours}_{\mathrm{FB}\alpha}$ TTA}, has the lowest average rank among the top four previous approaches. In particular we see that our method is a clear leader in the connectivity metric where Deep-Learning based matting methods have usually performed poorly. For comparison, Deep Image Matting~\cite{DeepImageMatting} ranks best among these methods for connectivity yet is ranked 9th overall, behind Closed-Form Matting~\cite{ClosedFormMattingPAMI} which ranks 7th, our method ranks 1st. We also submitted our model \textbf{$\text{Ours}_{\alpha}$ TTA} for to evaluate the effect of our foreground and background loss and fusion, and we found that the \textbf{$\text{Ours}_{\mathrm{FB}\alpha}$ TTA} performed best in all metrics except gradient. The full tables together with alpha matte predictions are available on \url{www.alphamatting.com}. 

\begin{table}[t]
\centering
\caption{Our scores in the \url{alphamatting.com} benchmark~\cite{alphamattingcom} together with the top-performing published methods. S, L, U denote the three trimap sizes, small, large and user, included in the benchmark. The scores denote the average rank of each method in these metrics across the 8 evaluation images.} \label{tab:alphamattingcom}
\begin{tabular}{lcccc<{\hspace{.8em}}cccc<{\hspace{.8em}}c<{\hspace{.8em}}c}
  \toprule
Method                                                & \multicolumn{4}{c}{Grad.}                                                                                                                            & \multicolumn{4}{c}{MSE}                                                                                                                            & SAD                                 & Conn.                               \\ 
                                                      & {\small overall}
                                                                                                                                                                                                             &
                                                                                                                                                                                                               S                                   & L                                   & U                                   & {\small overall}                             & S                                   & L                                 & U                                   & {\small overall}                             & {\small overall}                             \\  \midrule
\textbf{$\text{Ours}_{\mathrm{FB}\alpha}$ TTA}                                         & {\color[HTML]{FE0000} \textbf{1.4}} & {\color[HTML]{FE0000} \textbf{1.4}} & {\color[HTML]{FE0000} \textbf{1.3}} & {\color[HTML]{FE0000} \textbf{1.6}} & {\color[HTML]{FE0000} \textbf{1.7}} & {\color[HTML]{FE0000} \textbf{1.6}} & {\color[HTML]{FE0000} \textbf{1}} & {\color[HTML]{FE0000} \textbf{2.4}} & {\color[HTML]{FE0000} \textbf{1.9}} & {\color[HTML]{FE0000} \textbf{6.7}} \\

AdaMatting~\cite{DisentangledMatting}                 & 7.6                                 & {\color[HTML]{3531FF} 4.5}          & {\color[HTML]{3531FF} 5.3}          & 13                                  & {\color[HTML]{3166FF} 8}            & {\color[HTML]{3531FF} 5.6}          & {\color[HTML]{3531FF} 7.4}        & 10.9                                & {\color[HTML]{3531FF} 7}            & 19.5                                \\
SampleNet Matting~\cite{LearningBasedSamplingMatting} & 9.1                                 & 5.4                                 & 6.9                                 & 15                                  & 9                                   & 5.8                                 & 9.1                               & 12.1                                & 7.6                                 & 21.5                                \\
GCA Matting~\cite{GCAMatting}                         & {\color[HTML]{3531FF} 7.3}          & 7.3                                 & 6.1                                 & {\color[HTML]{3531FF} 8.6}          & 9.3                                 & 9.3                                 & 8.3                               & 10.5                                & 8.4                                 & 17.5                                \\
Deep Image Matting~\cite{DeepImageMatting}            & 17.6
                                                                                                                                                                                                              & 14.5                                & 14.3                                & 24                                  & 13                                  & 11.6                                & 11.8                              & 15.6                                & 10                                  & {\color[HTML]{3531FF} 16.5}         \\
\bottomrule
\end{tabular}
\end{table}

\section{Conclusion}
This paper presents a deep-learning method for simultaneously estimating the foreground, background and the alpha map from a single natural image and trimap. Our primary contribution to the alpha matting training regime is to use a batch-size of one. This, combined with longer training at 45 epochs and TTA, has a bigger impact than the choices of loss functions. %

Finally, our proposed solution for simultaneous $\alpha$, $\fgt$, $\bgt$ prediction, can achieve state of the art performance for all predictions, without increased computational or memory cost. Our reworked foreground dataset, new exclusion loss and fusion mechanism improve the final composite quality and can easily be incorporated in future works.

\bibliographystyle{splncs04}
\bibliography{longstrings,references}
\end{document}